\documentclass{article}

\usepackage[final]{neurips_2023}

\usepackage[utf8]{inputenc}
\usepackage[T1]{fontenc}
\usepackage{hyperref}
\usepackage{url}
\usepackage{booktabs}
\usepackage{amsmath,amssymb}
\usepackage{graphicx}
\usepackage{microtype}
\usepackage{xcolor}
\usepackage{tikz}
\usetikzlibrary{arrows.meta,positioning,fit,backgrounds}
\usepackage[font=small,skip=2.5pt]{caption}
\definecolor{rspmgreen}{HTML}{2CA02C}
\definecolor{rspmblue}{HTML}{4C72B0}
\definecolor{rspmorange}{HTML}{DD8452}

\title{Let Them Sleep: Recursive Self-Pruning Memory \\
for Long-Horizon LLM Agents}

\author{%
  \normalfont
Kargi Chauhan \\
  Pratibha Revankar \\
  Sadiba Nusrat Nur \\
  Kyle Montgomery \\
  Chenguang Wang \\[5pt]
  \textit{University of California, Santa Cruz}
}
\begin{document}

\maketitle

\begin{abstract}
Agents that run for a long time have a memory problem, and it is not the one
most systems optimize for. Storing every trace and retrieving the nearest
neighbors works early and quietly stops working later: the store fills up with
failures the agent then repeats, and with facts that were true fifty turns ago.
RSPM (Recursive Self-Pruning Memory) borrows the fix biology already found. At
intervals the agent sleeps. It clusters its recent failures, distills each
cluster into a short rule, and deletes the raw episodes it just learned from.
Rules decay unless something reinforces them, and rules that recur across many
contexts get promoted into general ones. We tested this on three settings with
little in common: 19{,}008 SWE-bench issues, temporal consistency across 12
conversational benchmarks, and long-horizon QA on LoCoMo. Performance rose by
7.6 to 11.9 points in each. More interestingly, three separately tuned
experiments landed on nearly the same equilibrium, roughly 93\% fewer retry
loops and roughly 30\% less stored memory, which suggests that knowing what to
forget is a general property of agent memory rather than a benchmark-specific
trick.
\end{abstract}

\section{Memory debt}

Long-running agents are now common. One works through GitHub issues at
SWE-bench scale~\citep{jimenez2024swebench}, another holds a conversation across
thirty sessions, another answers questions pointing back to an arbitrary turn
in dialogues of several hundred turns~\citep{locomo2024}. The usual design
assumption is that memory should only grow: keep every trace, retrieve the
closest matches~\citep{lewis2020rag}. Closest cousins either page memory across
tiers~\citep{packer2023memgpt}, decay unused traces~\citep{zhong2024memorybank},
or store lessons alongside experience rather than in place of
it~\citep{shinn2023reflexion,wang2024voyager}---none consolidate then delete.
In our experiments, the grow-only assumption broke down the same way in every
domain.

The store bloats first, since entries accumulate linearly while the useful
capacity of top-$k$ retrieval does not, so a naive conversational baseline ends
up holding 675 entries and surfacing five. Then the agent starts repeating
itself: a failed approach is stored verbatim, looks relevant later, and gets
tried again. We measure these retry loops at 5.9\% to 6.7\% of all attempts, in
software engineering and in QA alike. Finally, the store goes stale. A note
about a deprecated API, or a preference the user changed forty turns ago, still
carries full weight, and baseline accuracy slides from about 70\% to about 58\%
as a conversation runs past 100 turns. Biology settled this long ago: during
slow-wave sleep the hippocampus replays recent episodes while the neocortex
extracts the pattern, so the episode goes and the lesson
stays~\citep{stickgold2005sleep,mcclelland1995complementary}, and whatever is
not consolidated decays on the Ebbinghaus
curve~\citep{ebbinghaus1885gedachtnis}. RSPM writes those two observations down
as an algorithm.

\section{How RSPM works}

\begin{figure}[h]
\centering
\begin{tikzpicture}[
  font=\small,
  node distance=3.5mm,
  box/.style={draw, rounded corners=1.5pt, align=center, inner sep=3pt,
              minimum height=8mm, line width=0.5pt},
  arr/.style={-{Latex[length=1.6mm]}, line width=0.5pt}
]
\node[box, fill=rspmblue!10, draw=rspmblue] (ep) {episodic\\store};
\node[box, right=of ep, fill=black!4] (trig) {trigger\\{\scriptsize every 10 / failure spike}};
\node[box, right=of trig, fill=rspmorange!15, draw=rspmorange] (clu) {cluster\\failures};
\node[box, right=of clu, fill=rspmorange!15, draw=rspmorange] (dis) {distil\\1 rule};
\node[box, right=of dis, fill=rspmorange!15, draw=rspmorange] (pru) {prune\\episodes};
\node[box, right=of pru, fill=rspmgreen!12, draw=rspmgreen] (rul)
  {rules\\{\scriptsize L3 $\to$ L2 $\to$ L1}};
\node[box, right=of rul, fill=black!4] (dec) {decay\\{\scriptsize half-life $W$}};

\draw[arr] (ep) -- (trig);
\draw[arr] (trig) -- (clu);
\draw[arr] (clu) -- (dis);
\draw[arr] (dis) -- (pru);
\draw[arr] (pru) -- (rul);
\draw[arr] (rul) -- (dec);
\draw[arr, dashed] (pru.south) ++(0,0) -- ++(0,-4mm) -| (ep.south);
\draw[arr, dashed] (dec.north) -- ++(0,4mm) -| node[pos=0.25, above, font=\scriptsize]
  {retrieval reweighted by surviving rules} (ep.north);
\end{tikzpicture}
\caption{One sleep cycle. Raw episodes are clustered, compressed into a single
rule, and then deleted (dashed arrow back to the store). Surviving rules are
promoted upward when they generalize and decay away when nothing reinforces
them.}
\label{fig:arch}
\end{figure}
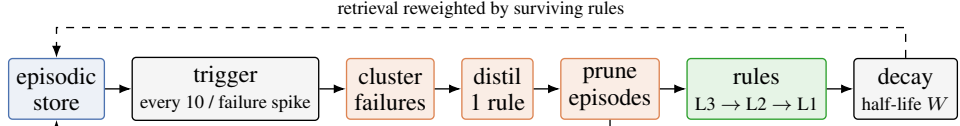

RSPM wraps an ordinary retrieval-augmented agent with three mechanisms
(Figure~\ref{fig:arch}). \textbf{Sleep cycles} fire every $S$ tasks ($S{=}10$)
and whenever the recent failure rate spikes. A cycle groups the recent
failures, turns each group into one compact constraint, then deletes the
episodes that produced it, so consolidation is paid for by pruning rather than
added on top of it. \textbf{Temporal decay} gives each unit of knowledge a
half-life, $P_{\text{eff}}(t) = P_0 \cdot \delta^{\lfloor (t-t_0)/W \rfloor}$,
and retires it below a threshold, so old lessons have to be re-earned.
\textbf{Promotion} lifts a constraint from instance level to category level
once it holds in three contexts, and to universal once it holds in five.

\section{What happened in three domains}
\label{sec:results}

We instantiated RSPM three times, each with its own notion of what makes an
entry prunable: failure clustering, temporal supersession, and failure
association.

\textbf{Software engineering.} On a difficulty-calibrated simulation over the
SWE-bench splits~\citep{jimenez2024swebench}, the solve rate on Full moves from
40.0\% to 49.1\%, a gain of 9.1 points ($p{=}0.0003$ over 10 seeds). The gain
is not uniform. It grows with difficulty, from 5.4 points on issues rated under
15 minutes to 13.2 on issues rated over four hours, which is what we would
expect if consolidation matters most under pressure. At 19{,}008 issues the
hierarchy yields 3{,}438 repo-specific constraints, 24 error-class rules and 18
universal ones, on 30.6\% fewer stored entries than the baseline.

\textbf{Temporal consistency.} Across 12 benchmarks and 1{,}363 items, RSPM
with an LLM conflict detector reaches 92.5\% temporal consistency. It beats
naive RAG at 80.6\%, which we expected, and it also beats an oracle given
ground-truth conflict labels at 88.4\%, which we did not. The oracle throws out
any document containing a superseded value, including the one announcing the
update, while the LLM detector reads the query first. The other surprise is
negative: keyword detection scores 79.5\%, slightly worse than doing nothing.

\textbf{Long-horizon QA.} On LoCoMo-10~\citep{locomo2024} (154 QA pairs per
conversation, 3{,}850 evaluations) RSPM reaches 72.8\% accuracy, 7.6 points
over baseline RAG ($p{=}0.0042$). The aggregate understates it: every baseline
decays as the conversation lengthens while RSPM does not, and its memory curve
flattens near 472 entries against 675. Ablating the sleep cycle costs more than
ablating anything else.

\section{What we take from this}

Pruning turned out to matter as much as retrieval. In all three settings, a
smaller curated memory beat a larger uncurated one, and the sleep cycle was the
mechanism we could least afford to lose. The convergence in the results is the part we find hardest to explain away: the
three studies were built at different times, on different data, with separately
tuned hyperparameters, none aimed at matching the others, yet they land on the
same two numbers. We read that as an equilibrium of the mechanism rather than
an artifact of tuning, and would like to be argued with on the point.

Forgetting also has a failure mode, and it is instructive. On one LoCoMo
conversation with a 35\% natural error rate, RSPM triggered 21 sleep cycles,
pruned 43\% of the store, and finished 19.5 points below baseline. Study~2
shows the same shape, with proactive archival hurting on datasets whose
questions ask about historical states. Both share an assumption we made too
casually, that a memory involved in a failure caused it. That holds when the
memory is stale and breaks when the task is simply hard. 

\newpage

{\small
\bibliographystyle{plainnat}
\bibliography{reference}
}

\end{document}